\documentclass{webofc}
\usepackage[varg]{txfonts}   

\usepackage{algorithm}
\usepackage{algorithmic}
\usepackage{natbib}
\usepackage{microtype}
\usepackage{graphicx}
\usepackage{subfigure}
\usepackage{booktabs}
\usepackage{bm}
\usepackage{multirow}
\usepackage{footnote}
\usepackage{multicol}
\usepackage{lineno}

\newcommand{\datatrain}[0]{\bm{x}_{\textrm{train}}}
\newcommand{\datavalid}[0]{\bm{x}_{\textrm{valid}}}
\newcommand{\multiML}[0]{\textsc{Multi-step ML}}
\newcommand{\KNAS}[0]{MSNAS}
\newcommand{\TASKONE}[0]{\textsc{Task~1}}
\newcommand{\TASKTWO}[0]{\textsc{Task~2}}

\begin{document}

\title{Event Classification with Multi-step Machine Learning}

\author{\firstname{Masahiko} \lastname{Saito}\inst{1,3}\fnsep\thanks{\email{saito@icepp.s.u-tokyo.ac.jp}} \and
        \firstname{Tomoe} \lastname{Kishimoto}\inst{1,3} \and
        \firstname{Yuya} \lastname{Kaneta}\inst{2} \and
        \firstname{Taichi} \lastname{Itoh}\inst{2} \and
        \firstname{Yoshiaki} \lastname{Umeda}\inst{2} \and
        \firstname{Junichi} \lastname{Tanaka}\inst{1,3}\fnsep\thanks{\email{jtanaka@icepp.s.u-tokyo.ac.jp}} \and
        \firstname{Yutaro} \lastname{Iiyama}\inst{1} \and
        \firstname{Ryu} \lastname{Sawada}\inst{1} \and
        \firstname{Koji} \lastname{Terashi}\inst{1}
}

\institute{International Center for Elementary Particle Physics, The University of Tokyo, 7-3-1 Hongo, Bunkyo, Tokyo, Japan
\and
           BrainPad Inc., 3-2-10 Shirokanedai, Minato, Tokyo, Japan
\and
           Institute for AI and Beyond, The University of Tokyo, 7-3-1 Hongo, Bunkyo, Tokyo, Japan
          }

\abstract{%
  The usefulness and value of \textsc{Multi-step} Machine Learning~(ML), 
  where a task is organized into connected sub-tasks with known intermediate inference goals, as opposed to a single large model learned end-to-end without intermediate sub-tasks,
  is presented.
  Pre-optimized ML models are connected and better performance is obtained by re-optimizing the connected one.
  The selection of an ML model from several small ML model candidates for each sub-task has been performed by using the idea based on Neural Architecture Search~(NAS).
  In this paper, Differentiable Architecture Search~(DARTS) and Single Path One-Shot NAS~(SPOS-NAS) are tested, where the construction of loss functions is improved to keep all ML models smoothly learning.
  Using DARTS and SPOS-NAS as an optimization and selection as well as the connections for multi-step machine learning systems,
  we find that (1) such a system can quickly and successfully select highly performant model combinations, and
  (2) the selected models are consistent with baseline algorithms, such as grid search, and their outputs are well controlled.
}

\maketitle

\section{Introduction}
\label{sec:intro}
\begin{figure}[ht]
\begin{center}
  {
    \subfigure[Single-step]{\label{fig:singleml}%
      \includegraphics[height=4.1cm]{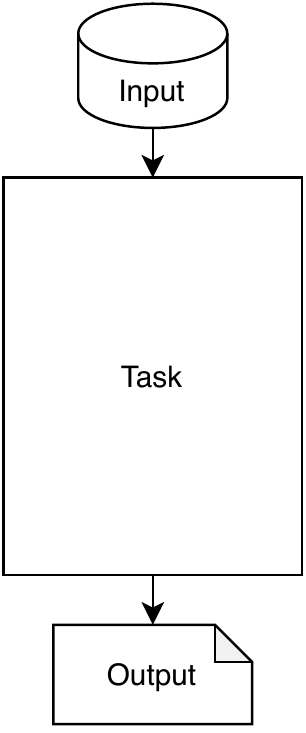}}%
    \qquad \qquad
    \subfigure[Multi-step]{\label{fig:multistep}%
      \includegraphics[height=4.1cm]{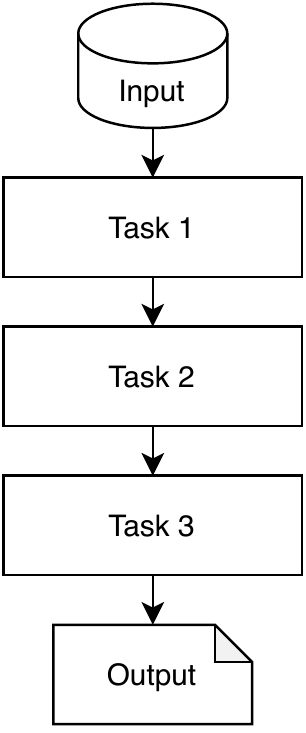}}
    \qquad \qquad
    \subfigure[Multi-step with model selection]{\label{fig:multiml-darts}%
      \includegraphics[height=4.1cm]{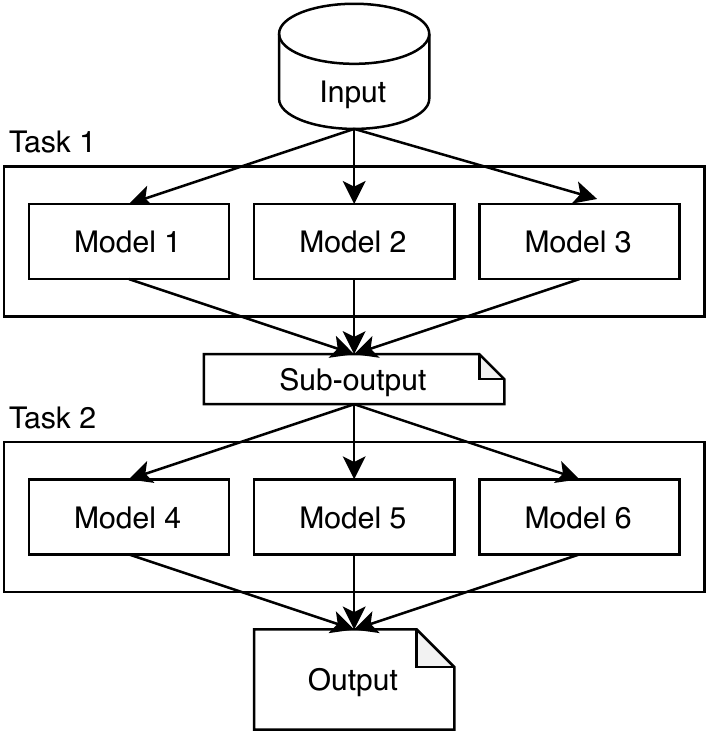}}
  }
  \caption{ML task flow}
\end{center}
\end{figure}

Machine learning~(ML), in particular, deep learning~(DL), has evolved rapidly due to the availability of huge computing power and big data
and has proven to be successful in many applications such as image classification, natural language translation, etc.
In most ML approaches, a single task with a large model learned end-to-end is defined and trained to solve a given problem~(see Fig.~\ref{fig:singleml}).
In most cases, this end-to-end approach provides state-of-the-art performance for a given problem in terms of precision and accuracy.
However, we will adopt a different approach, which can still give acceptable precision and accuracy for a given problem:
we connect some ML models, each of which can solve a part of a given problem as shown in Fig.~\ref{fig:multistep}.
We call it \multiML.
Moreover, in some of \multiML, we can assume that there are several different ML model candidates to solve the same sub-tasks as shown in Fig.~\ref{fig:multiml-darts}.
In this paper, ideas for the connecting of sub-tasks and their model selection are presented.

\paragraph{Multi-step ML}
For a given task, we break it into several sub-tasks with known intermediate inference goals, find optimal ML models for each sub-task, resulting in the best model chain.
From a different perspective, assuming that there are several tasks with well-defined or well-trained ML models, we solve a new task by combining them.
The common interesting point is that there are multiple sub-tasks for a given task.
ML models for each sub-task are relatively easy to build, or well-defined or well-trained ML models already exist.

Our approach may result in an interpretability versus accuracy trade-off when compared with end-to-end paradigms. The merits of our approach include:
\begin{itemize}
\item Domain knowledge is easily introduced into ML models for sub-tasks, and such ML models can be reused in other problems which involve common tasks,
\item Intermediate data, which is the output of sub-tasks, provide the information to understand the behavior of the ML models, which can lead to an explainability of ML.
\end{itemize}

The simplest way to connect ML models is just to give the output of an ML model to the input of a next ML model.
In this paper, we introduce more effective methods on the connection of ML models.

\paragraph{What type of problems can be useful for Multi-step ML?}
Sub-tasks have to be defined so that a given problem is required to be expressed as a set of sub-problems,
where such sub-problems should be recognized and their role, input and output are defined by a human.
Said in this exaggerated manner, this might correspond to problems found in any community.
A simple example is from some image classification problems: a sub-task to identify objects in an image and a sub-task to understand the context of the image
using the identified objects can be separated~\cite{9137382, kishimoto2021improvement}.
From the viewpoint of data flow, input data for \multiML\ are produced via several steps; in other words, there is a hierarchical structure in the input data.
For this type of data, we can solve the problem by defining sub-tasks for each step.
Practically, a problem to use data produced by a simulation of any model or theory is optimal
since the data could have a hierarchical structure or sub-tasks can be defined easily using supervised information.
This matches studies in science fields, for example, experimental particle physics, as shown later.

\paragraph{NAS to select MLs}
The situation shown in Fig.~\ref{fig:multiml-darts} might happen when there are different models requiring the optimization of ML model structures and hyperparameters.
We also expect that the choice of ML in the intermediate steps is not unique, that is, it will depend on the goal of a given task.
To select one of the models for each sub-task, in this paper, we use the idea of neural architecture search~(NAS). \\

We demonstrate the usefulness and value of \multiML\ on
(1) the re-optimization of model weights after sequentially connecting multiple ML models, and
(2) the selection of an ML model from multiple ML model candidates.
For the latter, we adopt the idea of the differentiable architecture search~(DARTS~\cite{liu2019darts}) and the single path one-shot neural architecture search~(SPOS-NAS~\cite{guo2020single})
to a task of particle physics, where two sub-tasks are defined to solve a given task.
This is called \textit{Model Selection with NAS~(MSNAS)}.

\paragraph{Particle Physics}
Experimental particle physics aims to understand the fundamental laws of nature and reveal unknown laws using a huge amounts of data.
In collider physics experiments, each event\footnote{
The term of ``event'' corresponds to the term of ``image'' in the image classification using for example CIFAR-10.}
of data is produced from collisions using a high energy accelerator.
An event has several particles, which are measured with detectors surrounding collision points.
The classification of events is quite important in collider physics data analysis, where interesting signal events are separated from background events.
ML has been used in collider physics research, for example, boosted decision trees~(BDT) for event classification~\cite{ROE2005577},
and DL for event classification~\cite{Baldi_2014}, jet imaging~\cite{Kasieczka_2019}, etc.

\section{Related Work}
\label{sec:related-work}
\multiML\ can be categorized into so-called \textit{Automated Machine Learning}~(AutoML), for example, Ref.~\cite{automl_book}, 
where the hyperparameter optimization, meta-learning, NAS, etc. are described and discussed.
The scope of AutoML is huge and continues to grow.
One of differences from AutoML is that in this paper we focus on the connection of multiple ML models, 
where not only a task but also sub-tasks are defined by humans because each sub-task has its own purpose.

NAS was introduced to automatically design a network architecture for a given task with the best performance and less human intervention.
The purpose of \multiML\ is different from that of NAS, however, methods developed for NAS can be applicable to \multiML.
Several survey documents of NAS are found, for example, in Refs.~\cite{ren2020comprehensive, elsken2019neural}: 
image classification~\cite{zoph2017neural, zoph2018learning, real2019regularized}, object detection~\cite{zoph2018learning, chen2019detnas}, etc.
In some NAS algorithms, large computational resources are required, for example, due to the discrete search space of the architecture with the reinforcement learning.
To overcome this issue, the idea of one-shot NAS~\cite{brock2017smash, pmlr-v80-bender18a} is promising: ENAS~\cite{pham2018efficient}, DARTS~\cite{liu2019darts, chen2019progressive, dong2019searching}, ProxylessNAS~\cite{cai2019proxylessnas}, SPOS-NAS~\cite{guo2020single}, SNAS~\cite{xie2020snas} and so on.
In DARTS, a differentiable calculation in the search space is introduced using the softmax function.
In SPOS-NAS, a supernet training and architecture search is decoupled by using the single path one-shot approach.
In NAS, a neural architecture is selected to achieve the best performance for a given task.
On the other hand, in \multiML, an ML model is selected by considering the performance of both a given task and its sub-tasks.

\section{Methods}
\label{sec:methods}
In this paper, the idea of DARTS~\cite{liu2019darts} and SPOS-NAS~\cite{guo2020single} is used to select one of the ML models.
One of the motivations of these algorithms based on NAS is to improve computing complexity.
By optimizing all model combinations simultaneously, instead of optimizing all model combinations separately, compute time reduces drastically because of avoiding repetitive model training.
In a method based on DARTS, a network consists of parallelly connected ML models. All model weights in the network are optimized simultaneously for each mini-batch.
In a method based on SPOS-NAS, a network consists of randomly sampled ML models. Model weights in the randomly selected model combination are optimized for each mini-batch.

Briefly summarizing these algorithms below, we explain what we have additionally done for connecting and selecting models.

\subsection{Based on DARTS}
DARTS represents a search architecture as a graph, where an edge corresponds to an operation.
Each edge has an architecture weight~($\alpha$), which is used to aggregate the outputs $o_i$ on the same path with a softmax function
$o = \sum_{i \in \mathrm{path}} \mathrm{softmax}(\alpha_i) \cdot o_i$.
After the training of architecture weights, operations that have maximum $\alpha$ among the same path are selected as a final operation set.

We use this idea for model selection.
The operations, which are represented as edges, are replaced with models for a sub-task.
The outputs of each model are built using architecture weight $\alpha$, called a model architecture weight hereafter,
$ y_{t} = \sum_{i \in \mathrm{models}} \mathrm{softmax}(\alpha_i) \cdot y_{t,i}$,
where $y_{t,i}$ is an output of $i$-th model for the $t$-th sub-task.
After the training of model architecture weights, the models that have maximum $\alpha$ among sub-tasks are selected as a final model set.

First, a pre-training is applied: every single model is individually trained using ground-truth data.
Second, model architecture weights are optimized following the DARTS optimization, where model weights~($w$) and the model architecture weights~($\alpha$) are optimized separately.
Model architecture weights are updated to be optimal for validation data, while model weights are fitted to training data.
Finally, with fixed models selected by model architecture weights, model weights of selected models are re-optimized for training data.
The algorithm is outlined in Appendix.~\ref{app:KNASalgorithms}.

A loss function used in DARTS for \KNAS\ is built using the loss functions of each sub-task:
\begin{align}
  \mathcal{L}(\bm{y}^\mathrm{true}, \bm{y}^\mathrm{pred}) &= \sum_{t \in \mathrm{task}} v_t \cdot \left( 
    \mathcal{L}_t (y_t^\mathrm{true}, y_t^\mathrm{pred})
    + \sum_{i \in \mathrm{models}} \epsilon \cdot \mathcal{L}_t (y^\mathrm{true}_t, y_{t,i}^\mathrm{pred})
  \right), \nonumber 
\end{align}
where $\mathcal{L}_t$, $y_t^\mathrm{true}$, and $y_t^\mathrm{pred}$ are a loss function, a ground truth, and a model prediction for the $t$-th task, respectively.
The first term is a loss function for aggregated outputs for each model.
This term is necessary for differentiably updating model architecture weights.
The second term is a loss function for each model.
Without the second term, outputs for each sub-task do not converge to the targets~$y^\mathrm{true}$, because there are degrees of freedom that can significantly change the output values of individual models while not changing the loss function values due to interference between models connected in parallel.
In this study, $\epsilon$ is fixed\footnote{Optimization of $\epsilon$, e.g. depending on $\alpha$ instead of a fixed value, is future work.} to 1, and task weights~($v_t$) are hyperparameters with a normalization of $\sum_{t \in \mathrm{task}} v_t = 1$, 
which was found to be reasonable values satisfying with both the validity of each model output and the performance of the whole task.

We use an Adam optimizer with a learning rate of $10^{-3}$ for pre-training, model architecture weight determination, and post-training.
The training is terminated after 100 epochs or if a valid loss does not decrease in 10~(20) epochs in pre/post-training~(model architecture weight determination).

\subsection{Based on SPOS-NAS}
In Ref.~\cite{guo2020single}, in the context of NAS, the supernet optimization~(weight optimization) and architecture search are decoupled.
Weights are optimized after selecting a single path of architectures with a uniform path sampling.
Then, the architecture search is performed using the evolutionary algorithm.

We use this idea for model selection.
Model weights in each sub-task are optimized after selecting a single path of models with a uniform sampling.
Then, the model search, where the best model is selected for each task, is performed using the grid search algorithm instead of the evolutionary algorithm since the number of models is small in this study.
The algorithm is outlined in Appendix.~\ref{app:KNASalgorithms}.

A loss function used in SPOS-NAS for \KNAS\ is built using the loss functions of each model:
\begin{equation}
  \mathcal{L}(\bm{y}^\mathrm{true}, \bm{y}^\mathrm{pred}) =\mathbb{E}_{i}\left[ \sum_{t \in \mathrm{task}} v_t \cdot \mathcal{L}_t (y^\mathrm{true}_t, y_{t,i}^\mathrm{pred}) \right], \nonumber
\end{equation}
where $i$-th models for the $t$-th sub-task are randomly sampled.
In this study, task weights~($v_t$) are hyperparameters like DARTS in \KNAS, with a normalization of $\sum_{t \in \mathrm{task}} v_t = 1$.
The optimizer used and the strategy of training termination are the same as the DARTS method.

\section{Experiments and Results}
\label{sec:exp-results}
A toy problem from experimental particle physics is prepared to prove the concept of \multiML.
All datasets have been generated by Monte-Carlo simulation.

\subsection{Problem and task}
The problem used as an experiment in this paper is the classification of particle origin: one is a Higgs boson~($H$) and the other is a $Z$ boson.
The main differences between the two particles are their mass~(125~GeV for $H$, 91~GeV for $Z$) and spin~(0 for $H$, 1 for $Z$).
Both particles promptly decay into a pair of $\tau$-leptons~($H/Z \rightarrow \tau^\pm \tau^\mp$) with some probability, and the $\tau$-leptons then decay into various particles, leaving an energy deposit in the detectors.
With the signature left in the detector, $\tau$-lepton candidates are reconstructed, then the particle origin is identified.
In this paper, we separate this problem into two parts and define sub-tasks:
the first task~(\TASKONE) is the energy calibration~(measurement) of a $\tau$-lepton candidate,
and the second one~(\TASKTWO) is the classification of $H$/$Z$ using a pair of $\tau$-lepton candidates where the input is the output from \TASKONE.

A loss function of \TASKONE\ is defined as mean squared errors of $\tau$-lepton momentum
\begin{equation}
  \mathcal{L}_1 = \frac{1}{N_{\mathrm{events}}} \sum_{i \in \mathrm{event}} \left(
    \| \bm{p}_{1, i}^{\mathrm{pred}} - \bm{p}_{1, i}^{\mathrm{true}} \|^2
  + \| \bm{p}_{2, i}^{\mathrm{pred}} - \bm{p}_{2, i}^{\mathrm{true}} \|^2 \right), \nonumber
\end{equation}
where $\bm{p_{1}}$~($\bm{p_{2}}$) is a leading~(sub-leading) $\tau$-lepton momentum.
For the \TASKTWO, a binary cross-entropy loss
\begin{equation}
  \mathcal{L}_2 = -y^\mathrm{(true)}\log y -(1-y^\mathrm{(true)})\log(1-y) \nonumber
\end{equation}
is considered.
For stable training, the output of \TASKTWO\ is defined as logits instead of probability~(i.e. sigmoid(logits)) and used instead of the formula above.
Output aggregation and loss function, therefore, are defined as
\begin{align}
  y &= \sum_{i \in \mathrm{models}} \mathrm{softmax}(\alpha_i) \cdot y_{i} \ \ \mathrm{(DARTS \ method)}, \nonumber \\
  \mathcal{L}_2 &= max(y, 0) - y \cdot y^\mathrm{(true)} + \log\left( 1 + e^{- \left| y \right| } \right). \nonumber
\end{align}

In our problem setting, the two loss functions cannot be treated as having equivalent statistics: the loss of \TASKONE\ is a $\chi^2$ assuming a momentum resolution of 1 GeV, while the loss of \TASKTWO\ can be regarded as a negative log-likelihood based on Bernoulli distribution.
To match the scale of two loss functions, the loss of \TASKONE\ is scaled by $10^{-4}$, i.e. the momentum is normalized by 100 GeV, in our experiments.

\subsection{Dataset}
The data was produced with particle physics simulation programs\footnote{
We use Monte Carlo~(MC) simulation to generate data, which is a set of events.
We generated MC events with Pythia8~\cite{Sj_strand_2015}
assuming 13~TeV proton-proton collisions like the Large Hadron Collider at CERN.
The detector response was simulated using Delphes~\cite{de_Favereau_2014} with the CMS parameterization (delphes\_card\_CMS\_PileUp.tcl.)
}.
We use only hadronic $\tau$-leptons that decay into hadrons, not electrons nor muons.
\TASKONE\ uses reconstructed-level jet 4-vectors and calorimeter/tracker information as input variables, then predicts truth-level $\tau$-lepton momentum.
Calorimeter/tracker information is given as 16x16 pixel images, which is expected to be used to estimate the momentum of neutrino from $\tau$-leptons to calibrate $\tau$-lepton momentum.
The input/output formats of \TASKONE\ and \TASKTWO\ and the pixel image examples are found in Appendix~\ref{app:dataset}.
The transverse momentum~$p_{\mathrm{T}}$ of \TASKONE\ and \TASKTWO\ used in ML models is normalized by $p_{\mathrm{T}} \leftarrow \log(0.1 + p_{\mathrm{T}} \mathrm{ (GeV)})$, to fit the values into a reasonable range for machine learning algorithms. 
We have 50,000 events for both $H$ and $Z$, where 60\% for training, 20\% for validation and 20\% for test.

\subsection{Models}
\label{sec:models}
Three kinds of models are prepared for each task. 
For \TASKONE, Multi-Layer Perceptron~(MLP), CNN and a linear transformation method, called a scale factor method~(\textsc{Sf} ) hereafter, are defined.
For \TASKTWO, MLP, Long Short-Term Memory~(LSTM), and a simple mass method~(\textsc{Mass}) are defined.
MLP, CNN and LSTM models are typical deep learning models, while \textsc{Sf} and \textsc{Mass} models are based on conventional methods used in collider particle physics. They are robust compared to deep learning models and are expected to be not the best models because of their simplicity.

MLP and CNN models for \TASKONE\ consist of two blocks: image feature extraction and correction factor evaluation~(see Appendix~\ref{app:task1model}).
The second block is designed to output a momentum residual like ResNet~\cite{7780459}.
CNN has a good domain bias for image recognition, while MLP does not.
The MLP model for \TASKONE\ is expected to be overfitted due to its large number of trainable parameters in this problem. 
A \textsc{Sf} model for \TASKONE\ applies a linear transformation~($f(x)=ax+b$) for each variable~($p_\mathrm{T}, \eta, \phi$)\footnote{A \textsc{Sf} model has six trainable parameters.}.

An MLP model for \TASKTWO\ is a simple deep neural network with three hidden layers with 32 nodes.
An LSTM model for \TASKTWO\ is built by three stacked LSTM modules with 32 hidden nodes.
Two $\tau$-leptons, ordering by jet $p_\mathrm{T}$, are sequentially fed to the LSTM module.
A \textsc{Mass} model for \TASKTWO\ calculates a system mass\footnote{The system mass $M$ is $\sqrt{(\sum_i E_i)^2 - (\sum_i p_{i,x})^2 - (\sum_i p_{i,y})^2 - (\sum_i p_{i,z})^2}$.} of two $\tau$-lepton candidates, then applies it to an MLP with 2 hidden layers with 64 nodes.

In all models above, ReLU is used as an activation function. The model hyperparameters, e.g. the number of layers, are determined by scanning them for each the single model.

In addition, two models: \textsc{Zeros}~(the output is always 0) and \textsc{Noise}~(Gaussian noise $\sim N(\mu=0, \sigma^2=1)$) are prepared as dummy models
and are used in the model selection studies. We expect that if DARTS works well, these models should not be selected.
On the other hand, SPOS-NAS cannot have dummy models since weights of models (MLP, etc.) are largely affected if such dummy models are included in a single-path.
Experiments on SPOS-NAS are performed for models not including these dummy models.

Before performing any studies, each model is pre-trained using ground-truth data from the simulation as explained in Section~\ref{sec:methods}.

\subsection{Results}

\subsubsection{Re-optimization of ML models used in multiple steps}
\begin{figure}[htbp]
  \begin{minipage}{0.65\linewidth}
    \centering
    \includegraphics[width=0.8\columnwidth]{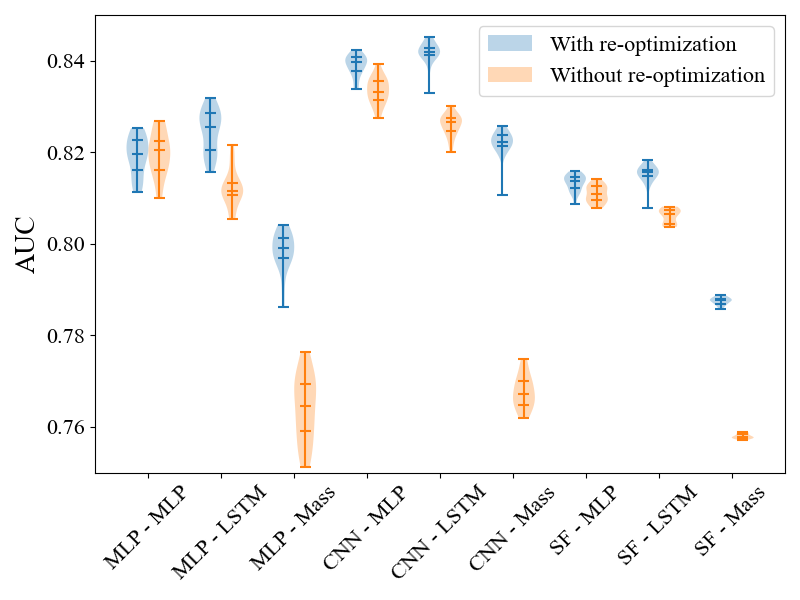}
  \end{minipage}
  \hspace{-15pt}
  \begin{minipage}{0.35\linewidth}
    \caption{
      AUC of \TASKTWO\ over 20 runs for all model pairs with~(blue) and without~(orange) re-optimization.
      Horizontal lines in the violin plot show the quantile values for 0, 0.25, 0.50, 0.75, and 1.
    }
    \label{fig:result_auc_grid_connection_separate}
  \end{minipage}
\end{figure}

We present the usefulness of the re-optimization of model weights with the sequentially connected with multiple trained ML models.
We execute experiments with two strategies:
\begin{description}
  \item [Without re-optimization]: Train a \TASKONE\ model, then train a \TASKTWO\ model using the outputs of the \TASKONE\ model.
  \item [With re-optimization]: Train \TASKONE\ and \TASKTWO\ models separately using ground-truth data from the simulation, then build a connected model and train the model.
\end{description}
A performance~(AUC) for \TASKTWO\ is measured for all model combination~(excluding dummy models) with and without the re-optimization as shown in Fig.~\ref{fig:result_auc_grid_connection_separate}, where experiments are executed 20 times with different random seeds for the same dataset.
For the re-optimization model, $v_1$ is set to zero\footnote{Two task weights~($v_1$ and $v_2$) are normalized, i.e. $v_2 = 1 - v_1$ in this experiment.}.
Re-optimization after pre-training improves the performance of the final task for any model pairs.
Pairs of (CNN, MLP) or (CNN, LSTM) have the highest AUC values in this experiment, and should be selected in \KNAS.

\begin{figure}[htbp]
  \centering
  {
    \subfigure[MSE of \TASKONE]{%
      \includegraphics[width=0.33\columnwidth]{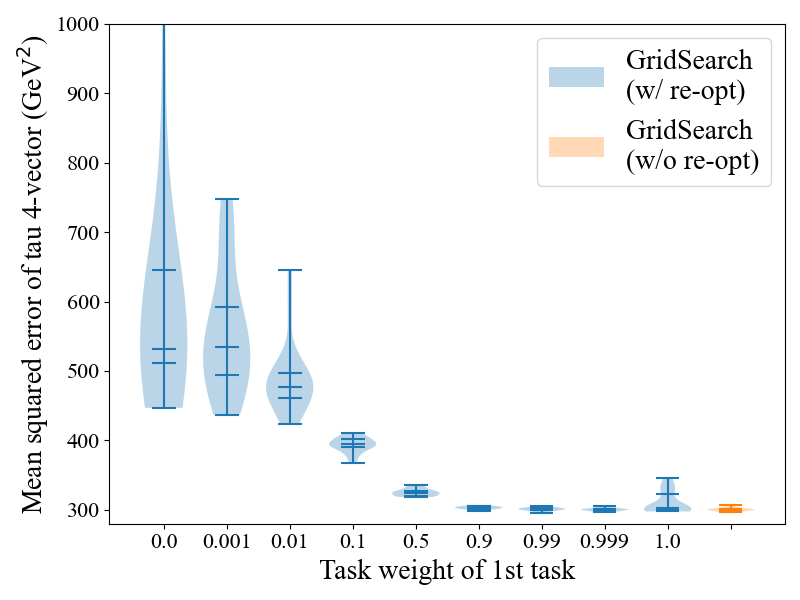}}%
    \subfigure[AUC of \TASKTWO]{%
      \includegraphics[width=0.33\columnwidth]{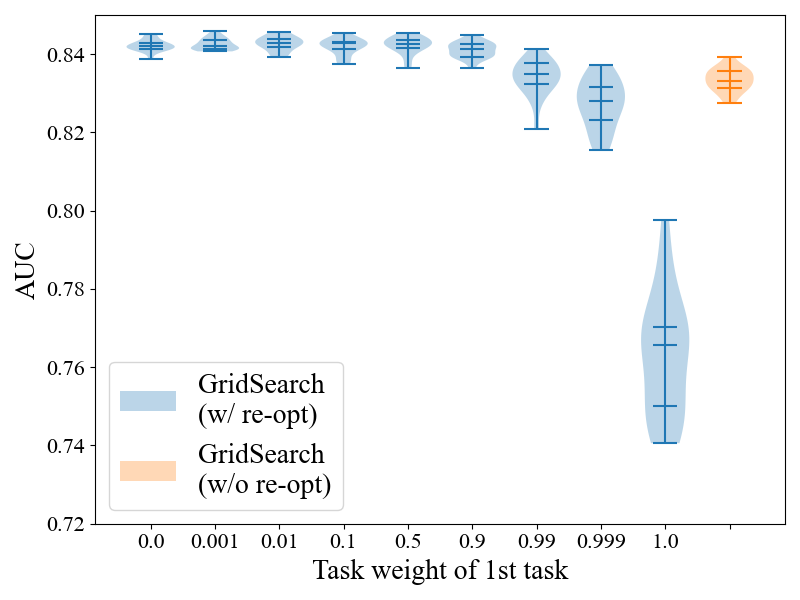}}%
    \subfigure[Fraction of events matching with Gaussian process predictions]{%
      \includegraphics[width=0.33\columnwidth]{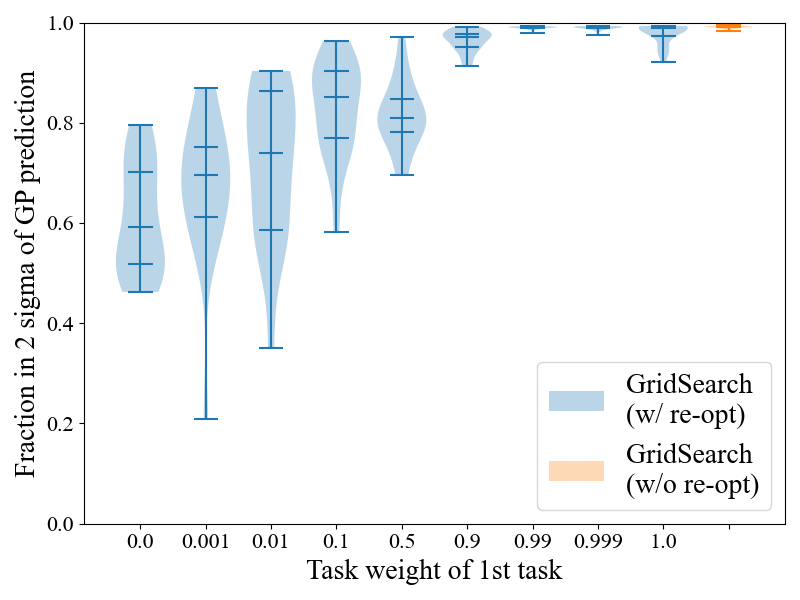}}
  }
  \caption{
    Mean squared error of \TASKONE\ (a) and AUC of \TASKTWO\ (b) as a function of \TASKONE\ weight $v_1$ for the re-optimization model~(blue) and without re-optimization~(orange).
    There are 20 runs for each point.
    The fraction of events where the model outputs are within the two sigma range of a Gaussian process prediction (c) as a function of \TASKONE\ weight $v_1$ for the re-optimization model~(blue) and without re-optimization~(orange).
  }
  \label{fig:result_loss_weights_grid}
\end{figure}

Figure~\ref{fig:result_loss_weights_grid}~(a) and (b) show loss-values of \TASKONE~(mean squared error) and AUC of \TASKTWO\ for the best model choices in each run as a function of \TASKONE\ weight $v_1$.
The larger the \TASKONE\ weight, the more that the \TASKONE\ loss value decreases, while the AUC values of \TASKTWO\ do not change significantly up to $v_1$ of 0.9.
The choice of task weights, $v_1 = 0.9$ in this case, improves explainability~(see the next section) while maintaining the performance of the final task.

\paragraph{Explainability}
By splitting a large problem into sub-tasks, \multiML\ is able to access intermediate states with readable forms.
To evaluate the interpretability of intermediate states, we measure the fraction of outliers of the \TASKONE\ output.
As a reference model that predicts robust and controlled outputs, we use a Gaussian process~(GP), which is a Bayesian machine learning technique predicting expected values with their uncertainties.
A Gaussian process is implemented using GPyTorch package, and it is trained using the same training data.
A validity of \TASKONE\ output is defined as the fraction of events where a model prediction of \TASKONE\ output is within the two sigma range predicted by Gaussian process\footnote{A consistency within 2$\sigma$ is independently evaluated for each scalar variables~($p_{\mathrm{T},1}, \eta_1, \phi_1, p_{\mathrm{T},2}, \eta_2, \phi_2$) for the simplicity.}.

A validity of \TASKONE\ is shown in Fig.~\ref{fig:result_loss_weights_grid} (c) as a function of \TASKONE\ weight~($v_1$).
Strong constraints on \TASKONE's loss result in similar predictions with a Gaussian process model.
Less constraints, on the other hand, give different predictions from the Gaussian process model over the uncertainties.
Such a prediction cannot be regarded as the one expected, e.g. particle momentum in this case.
A proper setting of task weights is required for the model to be explainable.

\subsubsection{Selection of an ML model from multiple ML models in each step}
We show results of the selection of an ML model using \KNAS.
To compare results with grid search and see the stability of the model choice,
we have 20 runs with the same dataset with different random seeds.
In no run were \textsc{Zeros} or \textsc{Noise} models selected by either \KNAS~(DARTS) or grid search.

\begin{figure}[htbp]
  \begin{center}
    {
      \subfigure[$\alpha$ for \TASKONE]{\label{fig:result_loss_weights_task1}%
        \includegraphics[width=0.42\columnwidth]{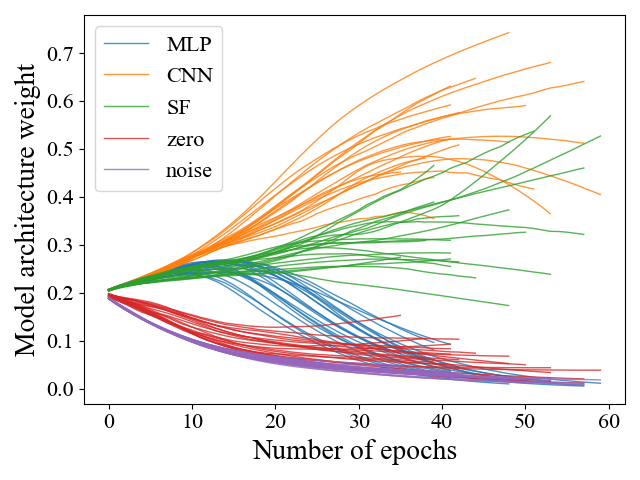}}%
      \subfigure[$\alpha$ for \TASKTWO]{\label{fig:result_loss_weights_task2}%
        \includegraphics[width=0.42\columnwidth]{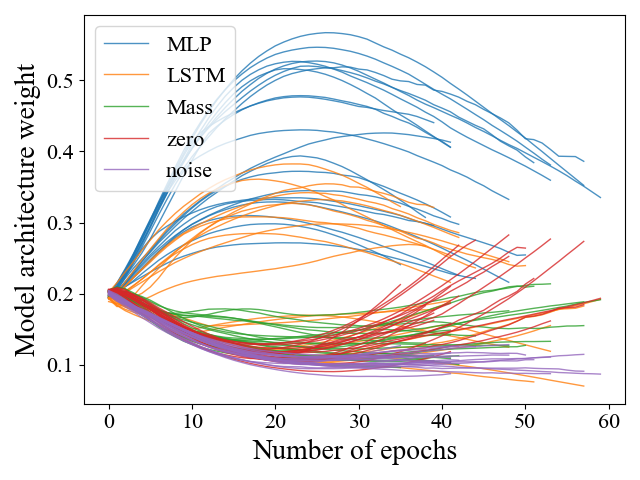}}
    }
    \caption{
      Model architecture weights (a) for \TASKONE\ and (b) for \TASKTWO\ in DARTS training step as a function of epoch with different colors for different models.
      All training is terminated by early stopping, where the training is forced to be terminated when the validation loss does not decrease in the last 20 epochs.
    }
    \label{fig:result_loss_weights}
  \end{center}
\end{figure}

The training process of the DARTS method is shown in Figs.~\ref{fig:result_loss_weights}, where model architecture weights~($\alpha$) in \TASKONE\ and \TASKTWO\ are shown as a function of the number of training epochs.
The behavior of $\alpha$ looks similar in all the 20 runs.
DARTS optimization, therefore, is robust for different initialization of each model weight.
Model architecture weights of \textsc{Zeros} and \textsc{Noise} models decrease as the training proceeds, while model architecture weights of other ML models~(MLP, CNN and LSTM) increases at the beginning of the training, as expected.

\begin{figure}[htbp]
  \centering
  \begin{minipage}{0.70\linewidth}
    \subfigure{
      \includegraphics[width=0.65\columnwidth]{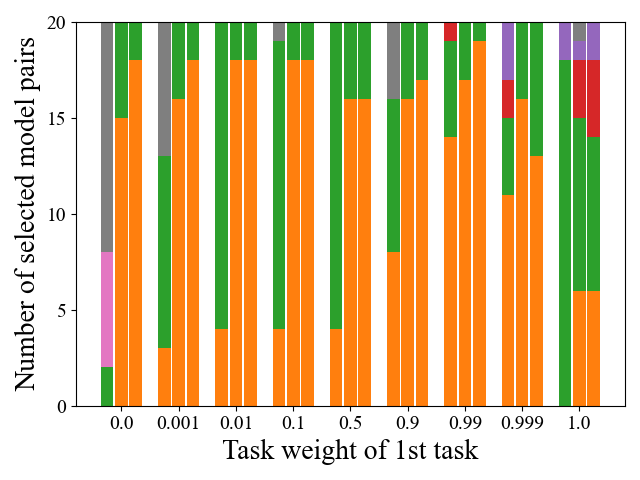}}%
      \hspace{-0pt}%
    \subfigure{
      \includegraphics[width=0.25\columnwidth, clip, trim=230 -20 50 10]{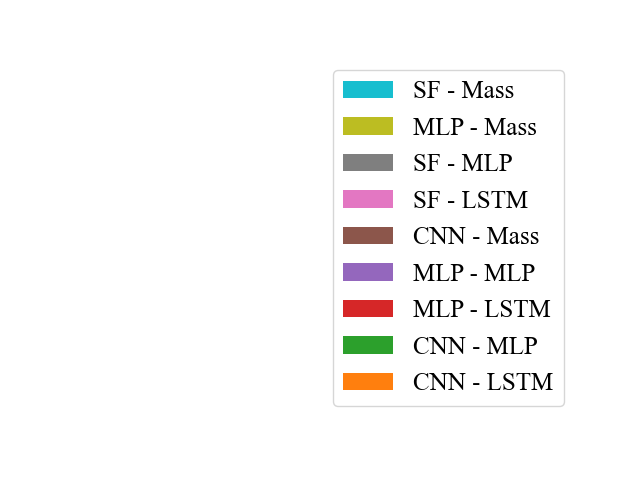}}
  \end{minipage}
  \hspace{-15pt}
  \begin{minipage}{0.30\linewidth}
    \caption{
      Selected model pairs in 20 runs as a function of \TASKONE\ weight $v_1$ for DARTS, SPOS-NAS, and grid search from the left.
      Model pairs are stacked in the order of performance for \TASKTWO\ measured in Fig.~\ref{fig:result_auc_grid_connection_separate}.
    }
    \label{fig:model_comb_grid_darts}
  \end{minipage}
\end{figure}

Figure~\ref{fig:model_comb_grid_darts} shows which models are selected in this experiment as a function of \TASKONE\ weight $v_1$.
Grid search and SPOS-NAS select (CNN, LSTM) and (CNN, MLP) pairs in this order, while DARTS selects these two model pairs but with different fractions.
Considering the difference of AUC between these pairs is small as shown in Fig.~\ref{fig:result_auc_grid_connection_separate},
DARTS can have a practical model selection power, but might have worse model selection ability than grid search and SPOS-NAS methods.

\begin{figure}[htbp]
  \begin{center}
    {
      \subfigure[Mean squared errors of \TASKONE]{%
        \includegraphics[width=0.33\columnwidth]{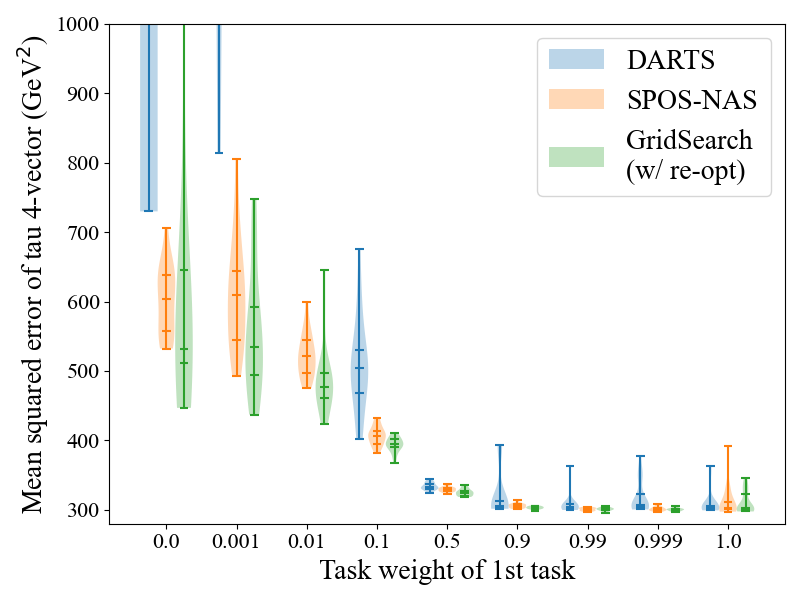}}%
      \subfigure[AUC of \TASKTWO]{%
        \includegraphics[width=0.33\columnwidth]{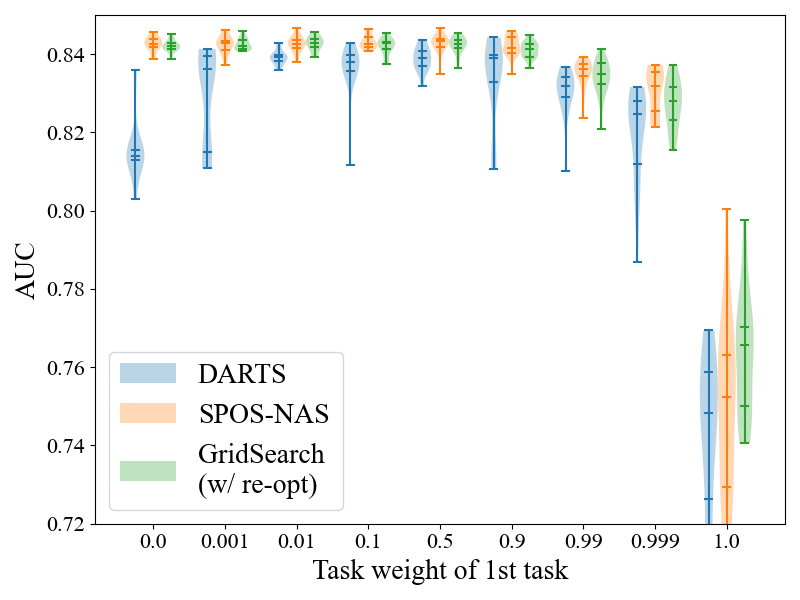}}
      \subfigure[Fraction of events matching with Gaussian process predictions]{%
        \includegraphics[width=0.33\columnwidth]{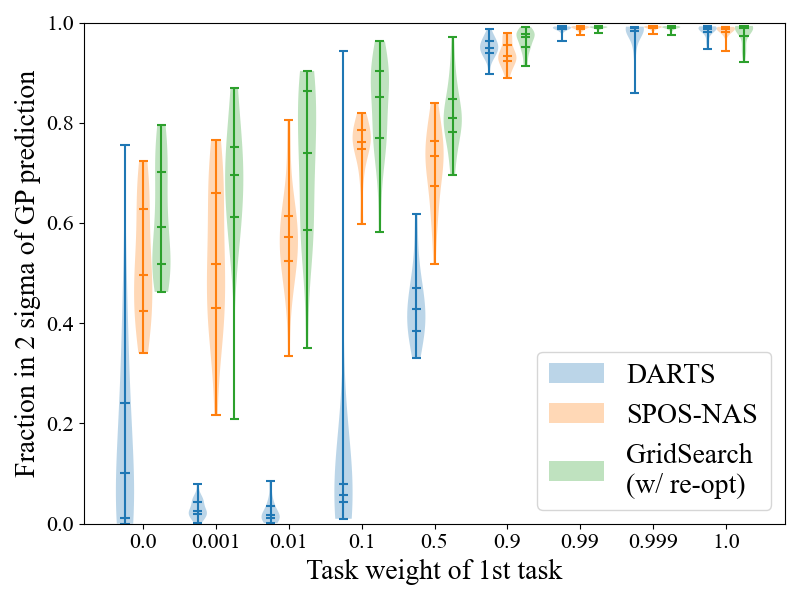}}%
    }
    \caption{
      (a) Mean squared errors of \TASKONE, (b) AUC of \TASKTWO\ and (c) the fraction of events matching with Gaussian process predictions for DARTS~(blue), SPOS-NAS~(orange) and grid search with re-optimization model~(green) as a function of a \TASKONE\ weight $v_1$.
      }
    \label{fig:result_auc_mse_task_weights_grid_darts}
  \end{center}
\end{figure}

Performance of \TASKONE~(MSE) and \TASKTWO~(AUC) is shown in Figs.~\ref{fig:result_auc_mse_task_weights_grid_darts}~(a) and (b) as a function of \TASKONE\ weight $v_1$.
As expected, as \TASKONE\ weight $v_1$ is larger, the performance of \TASKONE\ improves while that of \TASKTWO~(AUC) becomes worse.
There is, however, a moderate point where the performance for the both tasks does not change largely from the best point.
\KNAS\ gives a nearly optimal prediction for the last task with the intermediate data under control.
A validity check by Gaussian process prediction is performed as shown in Fig.~\ref{fig:result_auc_mse_task_weights_grid_darts}~(c).
Grid search and SPOS-NAS have similar performance, while DARTS has a quite different prediction compared to a Gaussian process at small $v_1$,
which will be investigated further.

\subsubsection{Scalability}
The compute complexity of a grid search is expressed by $O(\prod_{t \in \mathrm{tasks}} N_\mathrm{\mathrm{model, t}})$, i.e. $O(N_\mathrm{models}^{2})$ in our experiment, from the number of combinations for model pair.
On the other hand, the compute complexity of DARTS and SPOS-NAS is $O(N_\mathrm{tasks}N_\mathrm{models})$ because these models are trained simultaneously.

\begin{figure}[htbp]
  \begin{center}
    {
      \subfigure[Wall time]{\label{fig:walltime_nmodelsdep}%
        \includegraphics[width=0.46\columnwidth]{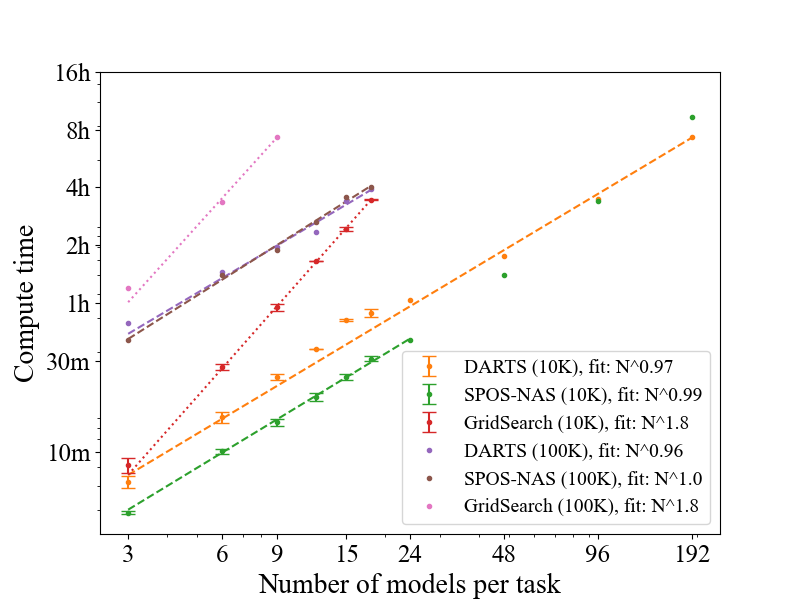}}%
      \subfigure[AUC]{\label{fig:auc_nmodelsdep}%
        \includegraphics[width=0.425\columnwidth]{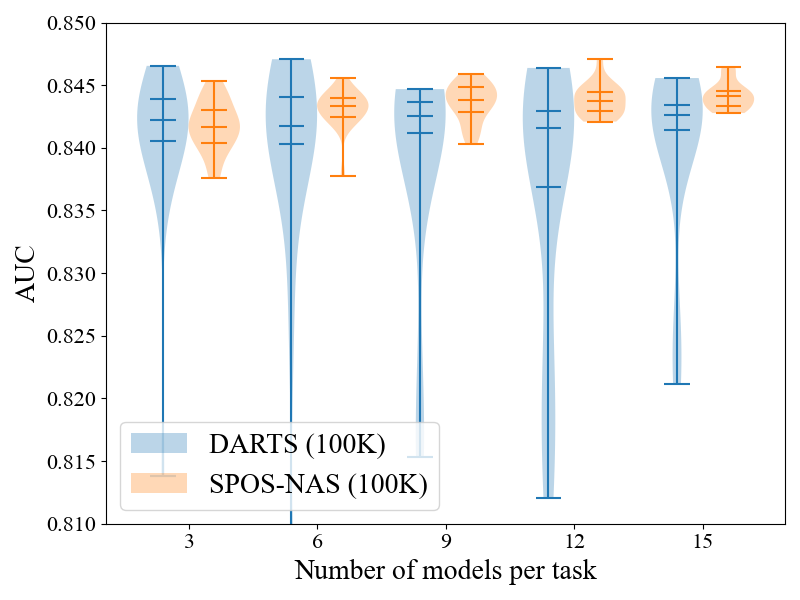}}%
    }
    \caption{
      (a) Wall time of \KNAS\ and grid search with two different-sized dataset 10K or 100K as a function of the number of models.
      NVIDIA Tesla A100 is used for all execution.
      DARTS and grid search are implemented using Tensorflow, and SPOS-NAS is implemented using PyTorch.
      Three~(one) independent runs are executed for small~(large) compute-time points drawn with~(without) error bars.
      Dashed~(\KNAS) and dotted lines~(grid search) show fitted lines with a function of $f(x) = C \cdot x^{a}$.
      For SPOS-NAS, the fitting is applied up to the first six bins, because the dependency of the number of models is not followed by a single power law.
      (b) AUC of \TASKTWO\ for a full dataset~(100K) as a function of the number of models, where 20 runs are executed for each point.
      }
    \label{fig:scalability_walltime}
  \end{center}
\end{figure}

We check the compute complexity as a function of the number of models.
In this experiment, the same models as explained in Section~\ref{sec:models} with dummy models excluded are used, i.e. the number of models for each task is three.
To measure the scalability with a large number of trainable models, the models above are replicated with different initialization, where model weights of replicated models are not shared.
Wall time for grid search, DARTS and SPOS-NAS is shown in Fig.~\ref{fig:scalability_walltime}~(a) as a function of the number of models per task, where two different sized datasets are used to measure the wall time within the reasonable execution time.
The dependency of the number of models follows our expectation: $O(N_\mathrm{models})$ for DARTS and SPOS-NAS and $O(N_\mathrm{models}^{2})$ for grid search.
SPOS-NAS uses grid search instead of a evolutionary algorithm after the one-shot NAS, resulting in changing the power law from $O(N_\mathrm{models})$ to $O(N_\mathrm{models}^{2})$.
This will be relaxed if a evolutionary algorithm is used.
Performance of model prediction of \TASKTWO\ is stable against the number of models as shown in Fig.~\ref{fig:scalability_walltime}~(b).
The DARTS and SPOS-NAS methods have good scalability for multiple-step machine learning problems.

\section{Discussion}
\label{sec:discussion}
We have performed \KNAS\ based on NAS techniques~(DARTS and SPOS-NAS) to connect and select ML models.
From the perspective of performance, an ensemble of several ML models or different weight initialization, where all trained MLs are used for inference, may improve the performance of the final prediction.
The purpose of \multiML, however, is different as we introduced in Section~\ref{sec:intro}:
sub-tasks are defined for a given a large problem, which might lead to making ML models simpler.
The number of model parameters can be reduced by selecting one of models with good performance, resulting in faster inference with lower computing resource requirements, e.g. memory.
Splitting into and defining sub-tasks instead of building a large single task increases the explainability, where all the outputs of sub-tasks can be accessed.
Moreover, it might be possible to know a proper domain for the problem from which a model is selected, e.g. if a LSTM model is selected, the problem has a strong correlation for a sequential structure.

Considering the demand from the machine learning community, hyperparameter optimization is also a hot topic.
\KNAS\ is able to select optimal hyperparameters as well as optimal models, by setting multiple models with different hyperparameters as model candidates in a task.
This method is scalable if each model has the small number of hyperparameter combination, e.g. $O(N_\mathrm{task} N_\mathrm{hp\_comb.})$
where each model has $N_\mathrm{hp\_comb.}$ hyperparameter combination.
However, it is not scalable if models have many types of hyperparameters, e.g. $O(N_\mathrm{task} N_\mathrm{hp}^{N_\mathrm{hp\_grid}})$
when each model has $N_\mathrm{hp}$ types of hyperparameters which have $N_\mathrm{hp\_grid}$ states.
Practically, the application for the hyperparameter scan requires our method to work well for discrete parameters. 
This is a future challenge.

The task weights~($v_i$) are treated as a hyperparameter in this study.
This parameter determines how much we respect the intermediate outputs.
On the other hand, we can treat them as floating parameters like the studies used in multi-task learning~\cite{8578879,pmlr-v80-chen18a,8954221,10.1007/978-3-030-01270-0_17,NEURIPS2018_432aca3a,NEURIPS2019_685bfde0}.
This is also a future challenge.

For the model selection, SPOS-NAS has nearly the same performance as grid search, but DARTS does not. This is under investigation.
Actually, our experiment is based on one specific problem, which is not familiar in the computing science field, so that we need more problems to test if our results are general.

For collider particle physics, the toy model used in this paper is a simplified model to demonstrate \KNAS\ methods.
In a more realistic case, there are more tasks to be considered, i.e. a tau identification task, or there is room to extend the object types and topology, e.g. $b$-jets, photon or leptons.
We plan to integrate such models and objects for \KNAS\ to be more practical in the future.

\section{Conclusions}
\label{sec:conclusion}
The usefulness and value of \multiML\ are presented in this paper.
The re-optimization after connecting multiple ML models gives better performance than the no re-optimization case.
The selection of a single ML model has been performed by using the idea of DARTS and SPOS-NAS,
where the construction of a loss function is improved to keep all ML models smoothly learning.
Using DARTS and SPOS-NAS as an optimization and selection as well as the connecting for multi-step machine learning systems, we find that (1) such system can quickly and successfully select highly performant model combinations, and (2) the selected models are consistent with baseline algorithms such as grid search and their outputs are well controlled.
Our idea has been tested for one specific problem so that we need more problems to test if our results are general.

\section*{Acknowledgments}
We thank Dr.~Michael Kagan (SLAC) and Dr.~Lukas Heinrich~(CERN) for the useful discussion and guidance for addressing ML techniques.
This research was partially supported by Institute for AI and Beyond of the University of Tokyo.

\section*{Code availability}
Our codes for the framework of \multiML~\cite{multiml_github} and for the implementation for this application~\cite{multiml_htautau_github} are available.

\bibliography{ms}

\clearpage

\appendix
\section{Details of \KNAS\ algorithm}
\label{app:KNASalgorithms}
\begin{algorithm}
  \footnotesize
  \caption{DARTS for Multi-step ML}
  \label{alg:alg_DARTS}
  \begin{multicols}{2}
  \begin{algorithmic}
  \STATE {\bfseries Notation} \\
      \hspace{5pt} $\bm{x}$: input/output data\\
      \hspace{5pt} $\{M_i^j\}$: $j$-th model for $i$-th task\\
      \hspace{5pt} $\bm{w}_i^j$: model weights of $M_i^j$\\
      \hspace{5pt} $\alpha_i^j$: model architecture weight of $M_i^j$\\
      \hspace{5pt} $\mathcal{L}^i$: loss function for $i$-th task\\
      \hspace{5pt} $\mathcal{L}$: loss function for multi-steps ML\\
  \smallskip
  \STATE{\bfseries \textsc{Pre-training}:}\\
  \FOR{each model~($M_i^j$)}
     \WHILE{not converged}
         \STATE Update $\bm{w}_i^j$ by descending $\nabla_{\bm{w}_i^j} \mathcal{L}^{i}(\bm{w}_i^j|\datatrain)$
     \ENDWHILE
  \ENDFOR
  
  \smallskip
  \STATE{\bfseries \textsc{Weights determination}:}\\
  \STATE Build a multi-step model parameterized by $\alpha_i^j$ for $M_i^j$
  \WHILE{not converged}
      \STATE Update $\alpha$ by descending $\nabla_{\alpha} \mathcal{L}(\bm{w}^{\ast}(\alpha), \alpha | \datavalid)$ \\
      \STATE Update $\bm{w}$ by descending $\nabla_{\bm{w}} \mathcal{L}(\bm{w}, \alpha | \datatrain )$
  \ENDWHILE
  \\
  \STATE $M_i^{\textrm{final}} \leftarrow M_i^{j_\mathrm{best}}, j_\mathrm{best} = \mathrm{arg~max}_{j \in \mathrm{models}} \alpha _i^j$\\
  \STATE Build a multi-step model using $\{M_i^{\textrm{final}}\}$\\

  \smallskip
  \STATE{\bfseries \textsc{Post-training}:}\\
  \WHILE{not converged}
      \STATE Update $\bm{w}_i^{\textrm{final}}$ by descending $\nabla_{\bm{w}_i^{j}} \mathcal{L}^{i}(\bm{w}_i^{j} | \datatrain)$
  \ENDWHILE

  \end{algorithmic}
  \end{multicols}
\end{algorithm}

\begin{algorithm}
  \footnotesize
  \caption{SPOS-NAS for Multi-step ML}
  \label{alg:alg_SPOSNAS}
  \begin{algorithmic}
  \STATE {\bfseries Same Notation, \textsc{Pre-training} and \textsc{Post-training} as DARTS} \\
  \smallskip

  \smallskip
  \STATE{\bfseries \textsc{Weights determination}:}\\
  \WHILE{not converged}
      \STATE Build a multi-step model using uniformly sampled models~($\{M_i^j\}$)
      \STATE Update $\bm{w}$ by descending $\nabla_{\bm{w}} \mathcal{L}(\bm{w} | \datatrain )$ for sampled model
  \ENDWHILE
  \\
  \WHILE{$\{M_i^j\}$ in all model combination}
    \IF {$\mathcal{L}(\{M_i^j\}) < \mathcal{L}(\{M_i^{\textrm{final}}\})$}
      \STATE $\{M_i^{\textrm{final}}\} \leftarrow \{M_i^j\}$
    \ENDIF
  \ENDWHILE
  \STATE Build a multi-step model using $\{M_i^{\textrm{final}}\}$\\

  \smallskip

  \end{algorithmic}
\end{algorithm}

\section{Details of Dataset}
\label{app:dataset}
\begin{figure}[htbp]
\begin{center}
  {
    \subfigure[Tracker]{\label{fig:input_task1_trk}%
      \includegraphics[width=0.30\columnwidth]{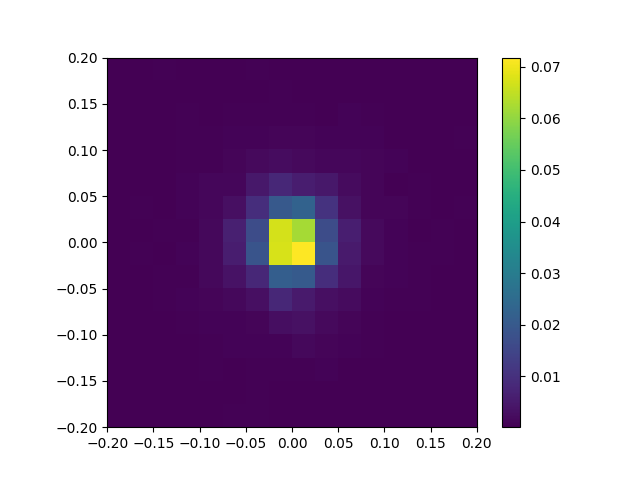}}%
    \subfigure[EM calo.]{\label{fig:input_task1_emcalo}%
      \includegraphics[width=0.30\columnwidth]{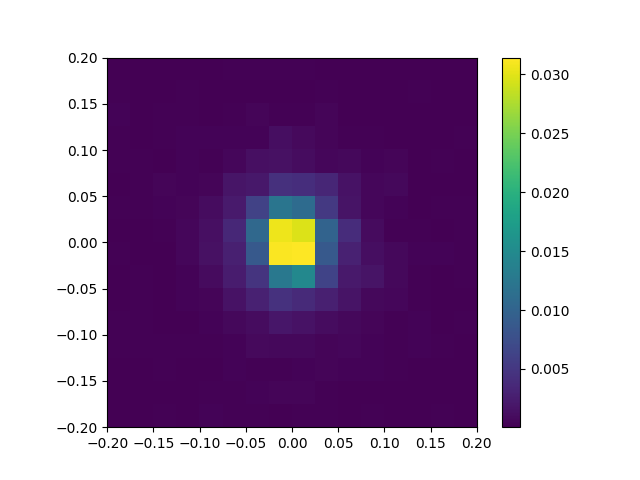}}
    \subfigure[Hadronic calo.]{\label{fig:input_task1_hadcalo}%
      \includegraphics[width=0.30\columnwidth]{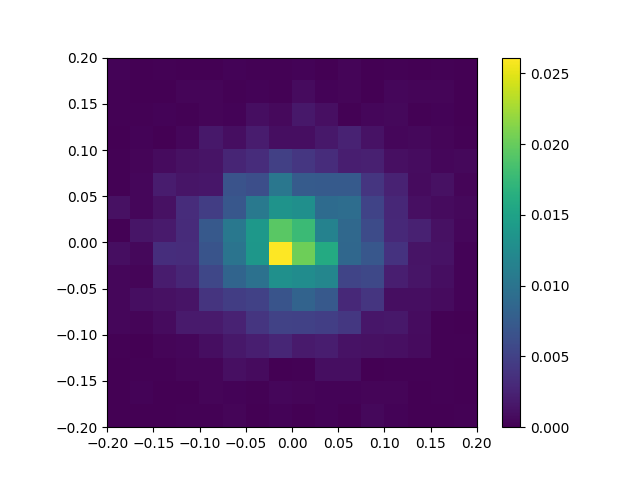}} \\
    \subfigure[Tracker]{\label{fig:input_task1_trk_1event}%
      \includegraphics[width=0.30\columnwidth]{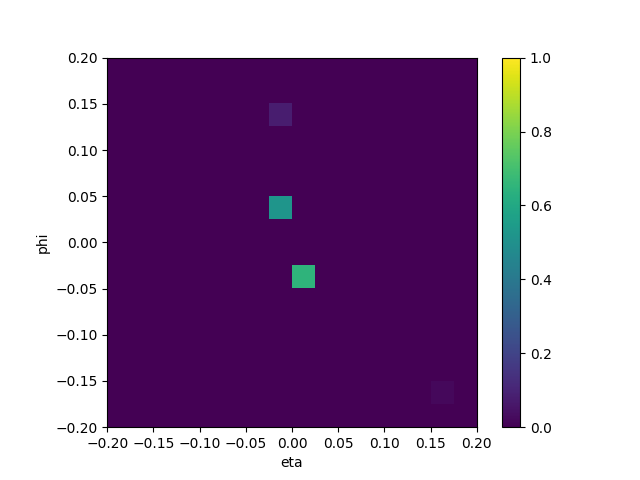}}%
    \subfigure[EM calo.]{\label{fig:input_task1_emcalo_1event}%
      \includegraphics[width=0.30\columnwidth]{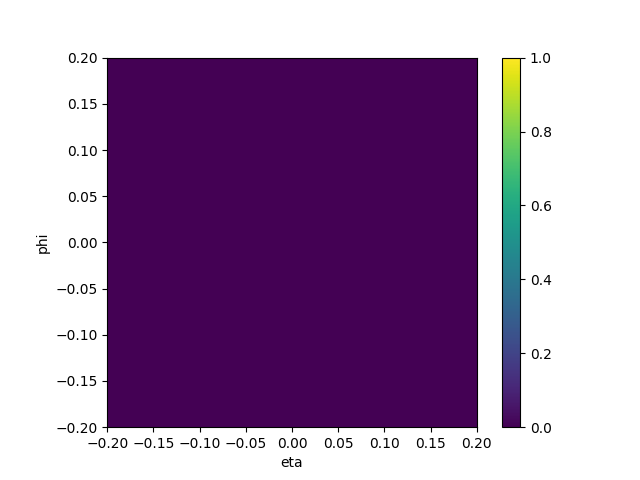}}
    \subfigure[Hadronic calo.]{\label{fig:input_task1_hadcalo_1event}%
      \includegraphics[width=0.30\columnwidth]{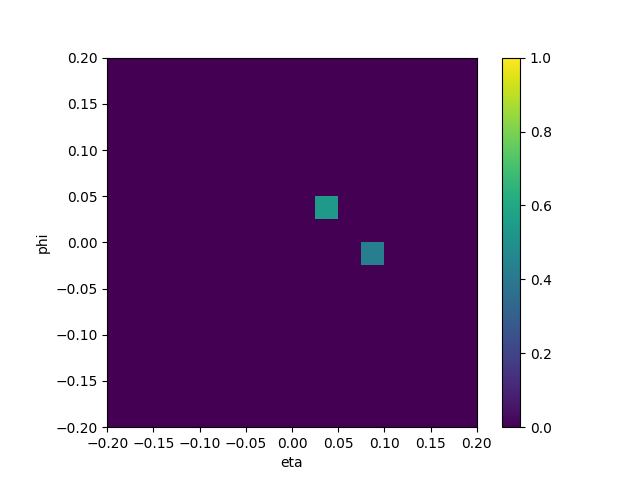}}
  }
  \caption{Distributions of input data in \TASKONE: (a-c) average over all events and (d-f) a single event.}
  \label{fig:input_task1_energymap}
\end{center}
\end{figure}

The input/output formats of \TASKONE\ and \TASKTWO\ is summarized in Tables~\ref{tab:dataset_input_task1} and \ref{tab:dataset_input_task2}, respectively.
Calorimeter/tracker information is given as 16x16 pixel images as shown in Figs.~\ref{fig:input_task1_energymap}.

\begin{savenotes}
\begin{table}
\begin{minipage}[t]{.45\textwidth}
  \caption{
    Input and output format for \TASKONE
  }
  \vskip -0.25in
  \label{tab:dataset_input_task1}
  \begin{center}
    \begin{footnotesize}
    \begin{sc}
    \begin{tabular}{ccc}
      \toprule
      & Name & Shape \\
      \midrule
      \multirow{8}{*}{Input} & Reconstructed jet & \multirow{2}{*}{(4, )} \\
      & 4 vector ($p_\mathrm{T}, \eta, \phi, m$) & \\
      \noalign{\smallskip}
      & Tracker $p_\mathrm{T}$ distribution & (16, 16) \\
      \noalign{\smallskip}
      & Electromagnetic~(EM) & \multirow{3}{*}{(16, 16)} \\
      & calorimeter & \\
      & $E_\mathrm{T}$ distribution & \\
      \noalign{\smallskip}
      & Hadronic calorimeter & \multirow{2}{*}{(16, 16)} \\
      & $E_\mathrm{T}$ distribution & \\
      \midrule
      Output & Tau 4 vector ($p_\mathrm{T}, \eta, \phi$)\footnote{The energy and 3-momentum $\vec{p}$ of a particle form a Lorentz vector~($E$, $p_x$, $p_y$, $p_z$) and can be converted into $p_\mathrm{T}=\sqrt{p_x^2+p_y^2}$, $\eta=-0.5\ln{(1-\cos\theta)/(1+\cos\theta)}$, $\cos\theta=p_z/|\vec{p}|$, $\phi=\texttt{atan2}(p_y,p_x)$, $m=\sqrt{E^2-p_x^2-p_y^2-p_z^2}$} & (3, ) \\
      \bottomrule
    \end{tabular}
    \end{sc}
    \end{footnotesize}
  \end{center}
\end{minipage}%
\hfill %
\begin{minipage}[t]{.45\textwidth}
  \caption{
    Input and output format for \TASKTWO
  }
  \vskip -0.25in
  \label{tab:dataset_input_task2}
  \begin{center}
    \begin{footnotesize}
    \begin{sc}
      \begin{tabular}{ccc}
      \toprule
      & Name & Shape \\
      \midrule
      Input & Tau 4 vector ($p_\mathrm{T}, \eta, \phi$) $\times 2$ & (6, ) \\
      \midrule
      Output & Higgs classification\footnote{This variable is close to $+\infty$ for $H$ but $-\infty$ for $Z$.} & (1, ) \\
      \bottomrule
      \end{tabular}
    \end{sc}
    \end{footnotesize}
    \end{center}
\end{minipage}
\vskip -0.1in
\end{table}
\end{savenotes}

\section{Details of \TASKONE\ Models}
\label{app:task1model}
\begin{figure}[htbp]
\begin{center}
  {
    \subfigure[MLP for \TASKONE]{\label{fig:model_task1_mlp}%
      \includegraphics[width=0.40\columnwidth]{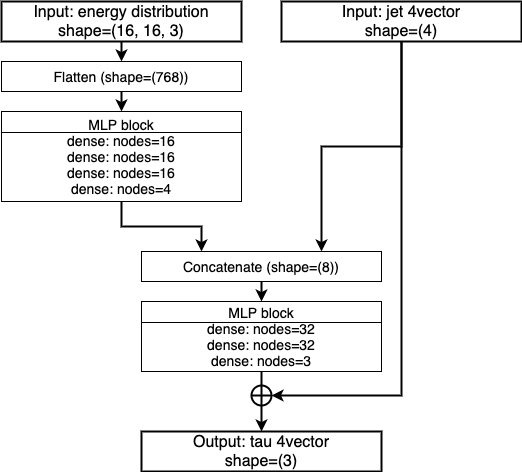}}%
    \qquad
    \subfigure[CNN for \TASKONE]{\label{fig:model_task1_conv2d}%
      \includegraphics[width=0.40\columnwidth]{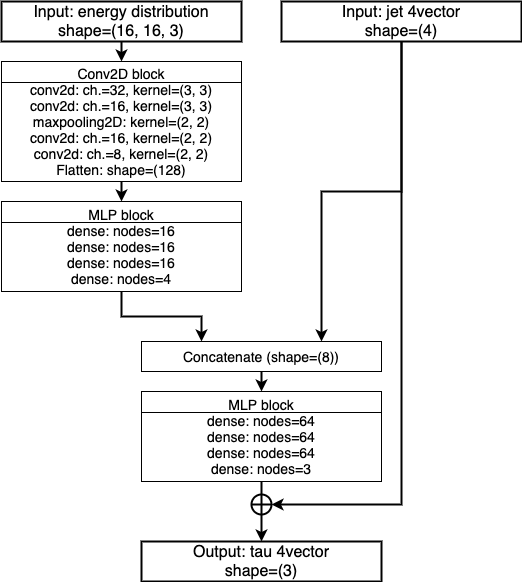}}
  }
  \caption{
    Models architecture of (a) MLP model and (b) CNN model used in \TASKONE.
    Both models consist of two blocks: image feature extraction and correction factor evaluation.
  }
  \label{fig:model_architecture_details}
\end{center}
\end{figure}

\end{document}